\documentclass[runningheads]{llncs}
\usepackage[T1]{fontenc}
\usepackage{booktabs}
\usepackage{float}
\usepackage{multirow}
\usepackage{color}
\usepackage{graphicx}
\usepackage{subfigure}
\usepackage{amsmath,amssymb,amsfonts}
\usepackage{color}
\usepackage{ulem}
\usepackage{hyperref}
\usepackage{soul}
\usepackage{ulem}
\usepackage[table]{xcolor}
\colorlet{shadecolor}{gray!20}
\usepackage{diagbox}

\hypersetup{
    colorlinks=true,
    linkcolor=blue,
    filecolor=magenta,      
    urlcolor=cyan,
    pdftitle={Overleaf Example},
    pdfpagemode=FullScreen,
    }
\begin{document}

\title{Distributionally Robust Optimization and 
Invariant Representation Learning for Addressing Subgroup Underrepresentation:\\ Mechanisms and Limitations}
%
%
\author{Nilesh Kumar$^*$, Ruby Shrestha$^*$, Zhiyuan Li \and Linwei Wang}
\institute{Rochester Institue of Technology, NY. USA}
%
\maketitle              
\vspace{-.3cm}
\begin{abstract}
Spurious correlation caused by subgroup underrepresentation has received increasing attention as a source of bias that can be perpetuated by deep neural networks (DNNs).
Distributionally robust optimization has shown success in addressing this bias, 
although the underlying working mechanism mostly relies on upweighting under-performing samples as surrogates for those underrepresented in data. 
At the same time,  
while invariant representation learning has been a powerful choice for removing nuisance sensitive features, 
it has been little considered in settings where spurious correlations are caused by significant underrepresentation of subgroups.
In this paper, we take the first step to
better \textit{understand} and \textit{improve} the mechanisms for debiasing spurious correlation due to subgroup underrepresentation in medical image classification. 
Through a comprehensive evaluation study, 
we first show that 1) generalized reweighting of under-performing samples can be problematic when bias is not the only cause
for poor performance, while 2) naive invariant representation learning suffers from spurious correlations itself. We then present 
a novel approach that leverages robust optimization 
to facilitate the learning of invariant representations at the presence of spurious correlations. 
Finetuned classifiers utilizing such representation 
demonstrated improved abilities to reduce subgroup performance disparity while maintaining high average and worst-group performance.        

\end{abstract}
\vspace{-.5cm}
\section{Introduction}

\def\thefootnote{*}\footnotetext{These authors contributed equally to this work}

As deep neural networks (DNNs) continue to demonstrate successes in various tasks \cite{he2016deep,rajpurkar2017chexnet}, their potential to generate and perpetuate bias also received growing attention \cite{oakden2020hidden,mccoy-etal-2019-right}. 
One source of biases being increasingly discussed is the presence of 
\textit{spurious correlation} due to \textit{subgroup underrepresentation,} 
where features irrelevant to a task happen to co-exist with a decision label in the majority of the samples (\textit{bias-aligning} samples) whereas samples that do not exhibit such correlation are underrepresented (\textit{bias-conflicting} samples). Examples may include the presence of treatment features in positive disease samples (\textit{e.g.}, the presence of drain as a treatment was observed 
in the majority of pneumothorax samples in CXR-14 chest X-rays \cite{oakden2020hidden}), or a higher prevalence of disease in certain demographic subgroups (\textit{e.g.}, malignant skin cancer was more commonly reported in lighter skin tones \cite{kinyanjui2020fairness}). With the standard
\textit{empirical risk minimization} (ERM), 
\cite{vapnik1991principles},
the DNN is trained to exploit such spurious correlation for reducing the \textit{average} loss of all training samples
\cite{mccoy-etal-2019-right}. Such 
DNN will struggle
with bias-conflicting subgroups.

\vspace{-0.06cm}
Most works have approached this bias 
from the lens of distributionally robust optimization (DRO), a classic optimization concept 
that focuses on minimizing worst-case losses over possible test distributions \cite{ben2013robust}. 
The recent integration of DRO into DNN was done by describing possible test distributions with importance sampling of the observed subgroups
\cite{sagawa2019distributionally}: the group DRO (GDRO) is then solved by iteratively optimizing the importance weights of all subgroups to maximize the DNN loss while optimizing the DNN to minimize such worst-case loss. This min-max formulation enables GDRO to pay equal or even greater attention to underrepresented subgroups, although it requires the subgroup labels to be known \textit{a priori}. To remove this need, various approaches have followed to identify bias-conflicting samples by, for instance, considering high-loss samples in a biased model \cite{pmlr-v139-liu21f,nam2020learning} or clustering the biased representations
\cite{sohoni2020no}. At the core of all of these approaches is \textit{a generalized re-weighting scheme to achieve DRO}, \textit{i.e.,} minimizing worst-case losses by upweighting the contribution from bias-conflicting subgroups.  
A fundamental working mechanism behind these approaches, however, assumes that high training loss of the ERM model can be used to determine the upweighting of bias-conflicting samples. While this was shown to be empirically effective on carefully designed benchmark datasets where spurious correlation is the dominating cause for poor performance, it is not clear how this working mechanism may generalize in real-world datasets where the high loss of a training sample may be due to other challenges such as the difficulty to extract discriminative semantics for certain disease categories.

\vspace{-0.06cm}
In parallel, \textit{invariant representation learning} has been a mainstream approach for removing confounding semantics not responsible for a task, especially for training DNNs that can adapt or generalize across domains \cite{https://doi.org/10.48550/arxiv.1505.07818}. It has also been widely used in \textit{fair representation learning} to obfuscate sensitive group attributes from latent representations \cite{deng2023fairness,https://doi.org/10.48550/arxiv.1511.00830,park2021learning,sarhan2020fairness,zhao2020training}. These works however either do not deal with spurious correlations due to underrepresented subgroups, 
deal with limited spurious correlation resulting from insubstantial bias ratio, or make the assumption that there exists a curated subset of data (for example, a training class) that does not exhibit subgroup underrepresentation. 
Within limited attempts of using invariant representations to handle spurious correlation, there are often specific assumptions of what may be the spurious factors (\textit{e.g.} texture of the image) and thus lacks generality  
\cite{wang2019learning}. To the best of our knowledge, there is very limited work on general purpose methods to learn representations invariant to spurious correlations \cite{pmlr-v162-zhang22z}.  
This leaves another intriguing open question: what may be the role and challenges of invariant representation learning in addressing spurious correlation caused by underrepresentation?

\vspace{-0.06cm}
In a nutshell, despite increasing interest in DNN fairness and bias \cite{10.1007/978-3-031-16431-6_70}, 
approaches to specifically address spurious correlation caused by subgroup representation have been limited in medical image classification \cite{jimenez2022detecting,sohoni2020no} and their working debiasing mechanisms and limitations are not well understood. 
In this paper, we take the first step to
better \textit{understand} and \textit{improve} the debiasing mechanism for addressing spurious correlation in medical image classification. 
To understand the potential mechanisms for debiasing and corresponding limitations (Section \ref{sec:evaluation}), we conduct a comprehensive evaluation study in which we examine
the aforementioned families of approaches to address spurious correlation on two skin lesion datasets. 
 Results suggested that 1) generalized reweighting based on under-performing samples 
 can be problematic in real-world datasets where bias is not the only cause for low performance, and 2) invariant representation learning suffers from spurious correlation and subgroup underrepresentation itself. 
 As a first step to address these observed limitations, 
 we further show that -- while GDRO in itself has little influence on representation learning -- its combination with domain adversarial loss can enable 
 invariant representation learning in the presence of spurious correlation (Section \ref{sec:newmethod}). Fine-tuning the classifier using such representation on balanced validation dataset, as inspired by deep feature reweighting (DFR) \cite{kirichenko2022last},
further improved the reduction of subgroup disparity while achieving high average performance.

\vspace{-0.3cm}
\section{Assessing Debiasing Mechanisms}
\label{sec:evaluation}

\subsection{Methodology}

\subsubsection{Data:} We consider two skin lesion datasets that exhibit different levels of complexity in the underlying tasks, biases, and the number of subgroups.

    \uline{\textit{ISIC}} \cite{codella2019skin}: The spurious correlation in ISIC 
    was originally caused by the presence of bandages in benign examples and the absence of bandages in malignant examples \cite{goel2020model}.
    Here we introduce a small subgroup of such malignant samples in the training to assess bias due to such underrepresentation. 
    This is achieved by 
       
       artificially adding bandages to malignant examples using the segmentation masks for bandages from \cite{rieger2020interpretations}.
    Table~\ref{tab:dataset-size} shows the distribution of training data. 
    
    \uline{\textit{Fitzpatrick}} \cite{groh2021evaluating}:
    As summarized in Table~\ref{tab:dataset-size}, in Fitzpatrick, there is a higher prevalence of malignant skin cancers in lighter skin tones (Fitzpatrick skin type 1-2) versus a higher  prevalence of non-neoplastic lesions in darker skin tones (Fitzpatrick skin type 5-6). This shift of label distribution among skin-type subgroups may create a bias of under-diagnosis of malignant skin cancers in individuals with darker skin tones.  
    To focus on this potential bias,
    we reduce overall skin-type imbalance by downsampling samples from skin types 1-4, while maintaining the same label distribution within each skin-type subgroup.

\begin{table*}[tbp]
\caption{Subgroup distribution in training. Bias-conflicting subgroups shaded gray.}
\label{tab:dataset-size}
\centering
\resizebox{\textwidth}{!}{%
\begin{tabular}{@{}llllll@{}}
\hline
\multicolumn{1}{c}{\multirow{1}{*}{\textbf{Dataset}}} & \multirow{1}{*}{} & 
\multicolumn{4}{c}{\textbf{Subgroup Distribution in Training Data}} \\ %
\hline
\multirow{1}{*}{\bf ISIC} && Benign-no bandage &  Benign-bandage &  Malignant-no bandage &  Malignant-bandage\\
&   {}
& 4843        & 4890        & 5205  
& \cellcolor{shadecolor}100          \\
\end{tabular} }
\\
\resizebox{\textwidth}{!}{%
\begin{tabular}{@{}llllllll@{}}
\hline
\multirow{1}{*}{\bf Fitzpatrick}  & \diagbox{Class}{Skin type} & 1 &  2 &  3 &  4 & 5 & 6\\
\hline
&    Benign         & 15.07\%        & 13.96\%        & 14.36\%  
&   13.20\%   & 10.37\% & 6.93\%   \\
&      Malignant      & 15.37\%        & 15.43\%        & 13.78\%  
&   10.82\%   & \cellcolor{shadecolor} 9.59\% &\cellcolor{shadecolor} 9.61\%   \\
& Non-neoplastic & \cellcolor{shadecolor}69.56\%        & \cellcolor{shadecolor}70.61\%        & 71.86\%  
&   75.98\%   & 80.04\% & 83.46\%   \\
\hline
\end{tabular}
}
\vspace{-.1cm}
\end{table*}

\vspace{-0.3cm}
\subsubsection{Models:} We consider three families of models to provide complete coverage of existing strategies
focused on DRO and/or representation learning. 

\uline{\textit{Generalized reweighting approaches:}}
Representing approaches using known subgroup labels, 
we consider simple importance weighting (with fixed subgroup weights) and GDRO \cite{sagawa2019distributionally} (with dynamic weights). 
Representing approaches with unknown subgroups, 
we consider JTT \cite{pmlr-v139-liu21f} where bias-conflicting samples are identified as under-performing samples in ERM, 
followed by upweighting. 

\par

\uline{\textit{Invariant representation learning:}} 
We consider a classic invariant representation learning strategy that removes domain information from the latent representation by a reversal gradient layer (known as domain adversarial training of neural networks/DANN) \cite{https://doi.org/10.48550/arxiv.1505.07818}. 
We use the bias-inducing factor as the domain, \textit{i.e.}
the presence and absence of bandage in ISIC and the skin types in Fitzpatrick.
We also consider the only existing general-purpose invariant representation learning approach reported for spurious correlation, where 
contrastive learning is used to push representations of samples of the same class closer (CnC) \cite{pmlr-v162-zhang22z}. 

\uline{\textit{Separated representation \& decision-boundary learning:}} 
We include deep feature reweighting (DFR) \cite{kirichenko2022last} that first obtains biased ERM representation, and then re-trains a classifier using such representations on a balanced validation set.

\vspace{-.4cm}
\subsubsection{Evaluation Metrics:} 
We consider three types of evaluation metrics. 
    
To \uline{\textit{evaluate bias in representations,}} 
we leverage self-organizing maps (SOM) \cite{kohonen1990self} to cluster and visualize the latent representations. We also adopt  
a quantitative metric of \textit{cluster purity} as a surrogate measure of biases in the latent representations. This is motivated by the concept that, in an ideal scenario, an unbiased representation should discard and thus not be separable by spuriously-correlated features.

To \uline{\textit{evaluate bias in the decision boundary,}} we consider two general types of quantitative metrics: 
1) test performance for the worst-performing subgroup, the anticipated bias-conflicting subgroups,
and averages across all subgroups; 
2) The performance disparity ($\Delta$) among all subgroups: in ISIC, this is measured as performance difference between worst-performing subgroups to best-performing subgroups ($\Delta_{best-worst}$) and to group-average performance ($\Delta_{avg-worst}$), respectively. In Fitzpatrick, because the number of subgroups is large ($3\times 6$), this is measured by $\Delta$ between best- and worst-performing subgroups in each class: malignant, non-neoplastic, and benign ($\Delta_{mn},\Delta_{nn},\Delta_{bg}$) as well as their average ($\Delta_{avg}$).

To \uline{\textit{assess the debiasing mechanisms,}}
for generalized reweighting with and without known subgroups, we track their importance weights, or the subgroups identified for up-weighting, respectively. For learning invariant representations, we track the accuracy of the domain classifier in removing spurious features.
\vspace{-.3cm}
\subsection{Experiments and Results}

\begin{table*}[t] \centering
\resizebox{\linewidth}{!}{
\begin{tabular}{@{\extracolsep{4pt}}lcccccccc@{\extracolsep{4pt}}}
\toprule
    & Avg & Worst & Malignant Bandage & $\Delta_{best-worst}$ & $\Delta_{avg-worst}$  \\

\midrule
ERM %
    & $ 0.803 \pm 0.005 $
    & $ 0.556 \pm 0.072 $ & $ 0.556 \pm 0.072 $ 
    & $   0.443 \pm 0.072 $ & $ 0.246  \pm 0.066 $  \\

Important weighting   %

    & $ 0.793 \pm 0.011 $
    & $ 0.593 \pm 0.040 $ & $ 0.593 \pm 0.040 $ & $ 0.406 \pm 0.040 $ & $ 0.200 \pm 0.030 $ \\


JTT %
    & $ 0.790 \pm 0.011 $
    & $ 0.550 \pm 0.051 $ & $ 0.550 \pm 0.051 $ & $ 0.450 \pm 0.051 $ & $ 0.240 \pm 0.041$ \\

DANN %
    & $ 0.843 \pm 0.005 $
    & $ 0.730 \pm 0.017 $ & $ 0.730 \pm 0.017 $ & $ 0.266 \pm 0.023 $ & $ 0.113 \pm 0.011 $ \\

DFR %
    & $ \cellcolor{shadecolor}0.871 \pm 0.004 $
    & $ 0.758 \pm 0.008 $ & $ \cellcolor{shadecolor}0.865 \pm 0.010 $ & $ 0.232 \pm 0.010 $ & $ 0.112 \pm 0.010 $ \\

GDRO 
    & $ 0.846 \pm 0.011 $
    & $ 0.760 \pm 0.017 $ & $ 0.763 \pm 0.015 $ & $ 0.240 \pm 0.017 $ & $ 0.086 \pm 0.005 $ \\

GDRO with group adjustment %
    & $ \mathbf{0.866 \pm 0.005} $
    & $ 0.786 \pm 0.011 $ & $ 0.803 \pm 0.020 $  & $0.213 \pm 0.011 $ & $ 0.080 \pm 0.005 $ \\

Proposed method %
    & $ \mathbf{0.863 \pm 0.005} $
    & $ \cellcolor{shadecolor}0.800 \pm 0.000 $ & $ \mathbf{0.813 \pm 0.023} $ & $ \cellcolor{shadecolor}0.190 \pm 0.000 $ &
    $ \cellcolor{shadecolor}0.063 \pm 0.005 $ \\

\bottomrule
\end{tabular}
}
\caption{\label{tab:all_methods_results}
Accuracy of benign \textit{vs.} malignant classification on ISIC. 
Malignant samples with a bandage is the underrepresented bias-conflicting group as expected from Table ~\ref{tab:dataset-size}. Columns 2-4 list accuracy of averaged and individual subgroups. Columns 5-6 measure the performance disparity among subgroups. Best performance is shaded gray, with bolded performance closely behind.}

\vspace{-.2cm}
\end{table*}

Following convention \cite{sagawa2019distributionally,pmlr-v162-zhang22z}, we used ResNet-50 from torch-vision with pre-trained weights from Imagenet. We used held-out validation set to choose the best model 
unless the method is using the validation set for fine-tuning (\textit{e.g.}, DFR).

\begin{figure}[!tb]
    \centering
    \includegraphics[width=1\textwidth,scale=0.5]{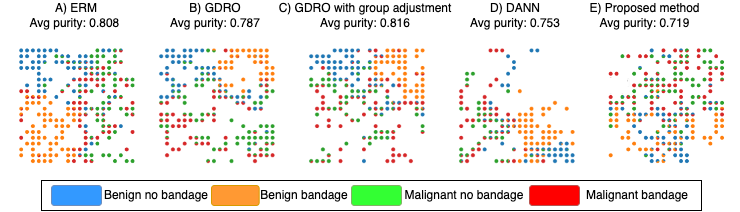}
    \caption{SOM plots and their averaged subgroup purities.
    }
    \label{fig:som_plots}
\end{figure}

\vspace{-0.3cm}
\subsubsection{ISIC:} As summarized in Table~\ref{tab:all_methods_results}, with ERM, the DNN suffered from a significant performance drop for the bias-conflicting subgroup (malignant samples with bandage). This bias is also evident in the SOM (Fig~\ref{fig:som_plots}A), which shows clearly separated subgroups and a relatively high purity score. 

\begin{figure}[t]
    \centering
    \includegraphics[width=1\textwidth,scale=0.5]{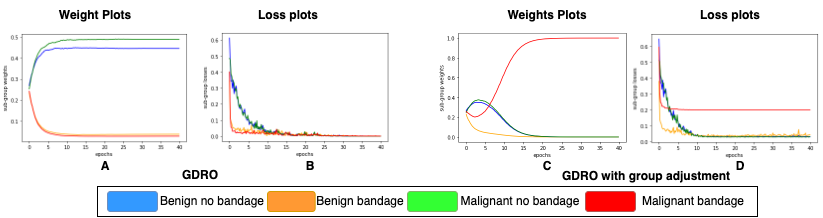}  \vspace{-.5cm}
    \caption{ \vspace{-.5cm}Weight and loss plots for GDRO with and without group adjustment.}
    \label{fig:weights_loss_plots_gdro}
\end{figure}

\uline{Generalized reweighting approaches:} Simple importance weighting (to equally weight all subgroups) improved the test accuracy on the bias-conflicting group but to a limited extent. 
GDRO more significantly improved the test performance of the bias-conflicting subgroup.
Interestingly, a closer inspection on the subgroup weights obtained by the GDRO (Figure.~\ref{fig:weights_loss_plots_gdro}A) showed  that GDRO actually upweighted the two non-bandage subgroups. 
Evidently, with group-based average loss, 
the underrepresented group is already upweighted in GDRO which  
quickly resulted in a lower loss at the early stage of training. As such,  
the other two non-bandage subgroups unexpectedly became the higher-loss subgroups 
(Figure.~\ref{fig:weights_loss_plots_gdro}B)
and were upweighted. The separation of subgroups in the SOM remained similar to ERM with slightly reduced purity (Fig~\ref{fig:som_plots}B).

Facing this unexpected working mechanism of GDRO, 
we further experimented with a version of GDRO with \textit{group adjustment} \cite{sagawa2019distributionally}: this adds an additional term of $\frac{1}{\sqrt{N_s}}$ to the subgroup loss where $N_s$ is the size of a subgroup. 
As expected, this term dominates the loss used to optimize the weights (Figure.~\ref{fig:weights_loss_plots_gdro}D) 
and
resulted in a heavy upweighting of the bias-conflicting group (Figure.~\ref{fig:weights_loss_plots_gdro}C). This resulted in further improvement on this subgroup. Clustering of the latent representation and its purity however remained similar to the ERM (Fig~\ref{fig:som_plots}C).

These results revealed that 
 the use of group adjustment played a significant role in the weighting mechanism of GDRO. Without this, the upweighting was influenced by other factors contributing to low performance and the overall results appeared to be adversely affected. 

\begin{figure}[t]
    \centering
    \includegraphics[width=0.8\textwidth,scale=0.5]{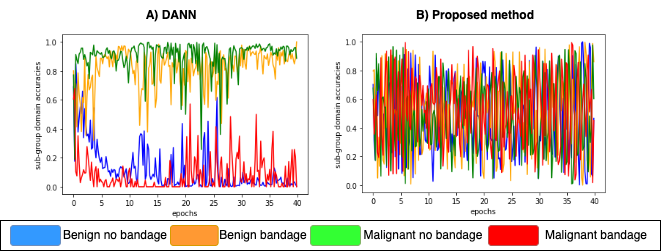}
    \caption{Subgroup accuracy of domain classifier in A) DANN and B) as proposed.}
    \label{fig:domain_accs}
\end{figure}

\uline{Invariant representation approaches:} 
DANN was trained with upsampling of the
underrepresented subgroup. 
Its performance was better than simple importance weighting, 
but inferior to GDRO. 
A closer look at the obtained representation clusters (Fig~\ref{fig:som_plots}D) and the behaviour of the domain classifier (Figure~\ref{fig:domain_accs}A) 
suggested that DANN was not successful in removing bandage information from the bias-aligning subgroup (benign with bandage). 
This failure suggested that domain-invariant representation learning also suffers from spurious correlations.

\uline{Approaches without subgroup labels:} For the remaining models that do not require the use of subgroup labels, the performance was suboptimal. JTT showed minor improvement over ERM, while CnC under-performed than ERM (results thus not included). Common to both approaches was the reliance on a biased ERM model to identify bias-conflicting examples. A closer inspection showed that, among the under-performing samples selected from the biased ERM, 
only 1.05-2.75\% belonged to the bias-conflicting subgroup. This  explained why the subsequent learning approaches may not be successful. 
Finally, DFR achieved an impressive average performance as well as on the bias-conflicting subgroup, albeit at the expense of creating a new worst-performing subgroup.

\vspace{-.3cm}
\subsubsection{Fitzpatrick:} 

Table \ref{tab:all_methods_results_fitz} shows the average AUC, worst-AUC among all 18 subgroups, and AUCs for four selected subgroups that are most indicative of the bias present: 
with ERM, 
the subgroups with darker skin tones are disadvantaged in malignant cancer detection (MN-56, AUC of 0.81 \textit{vs.}, 0.88 for MN-12), whereas the subgroups with lighter skin tones are disadvantaged in non-neoplastic lesion detection (NN-12, AUC of 0.78 \textit{vs.}, 0.86 for NN-56). 
This disparity in performance is perfectly associated with 
the subgroup label distributional shifts shown in Table \ref{tab:dataset-size}, an intriguing non-artificial bias that may be challenging to address.

The trend of performance of all models was similar to that observed in ISIC with some distinction. JTT and CnC continued to under-perform and their results are not included. 
Unlike on ISIC, here both DANN and DFR obtained negligible or even negative gains over ERM. 
In comparison,
various reweighting strategies 
were all effective in addressing the bias against MN-56, albeit  
1) at the expense of decreased average AUC, 
and 2) ineffective in addressing the bias against NN-12. 
The measures of performance disparity within and averaged across classes (Table \ref{tab:all_methods_results_fitz_delta}) suggested the same trend. 

Note that the worst-performance subgroups from all models were 
different from the bias-conflicting subgroups, 
which may be why this particular bias was challenging to remove.

\vspace{-0.3cm}
\section{Improving the Debiasing of Spurious Correlations}
\vspace{-.1cm}
\label{sec:newmethod}

\subsubsection{Methodology:}

The investigations in Section \ref{sec:evaluation} provided two important findings. First, the reliance on low training performance for identifying or upweighting bias-conflicting samples may face challenges where bias is not the only cause for under-performing samples. 
Second, while invariant representation in concept appears a natural candidate for removing confounders, its learning ironically also suffers from spurious correlation caused by underrepresentation.
This raises a critical question: is our best hope upweighting known bias-conflicting subgroups?

We re-examine DANN's failure to confuse the domain classifier on the bias-aligned subgroups (Fig.~\ref{fig:domain_accs}A). A possible explanation is that 
the main task classifier was exploiting spurious correlations and overpowers the domain classifier. 
Based on this, we hypothesize that a robust main classifier with a reduced tendency to exploit spurious correlation may better support DANN to learn representations invariant to spurious correlations. 
To this end, we propose to optimize the main classifier in DANN with GDRO and, with the learned invariant representation, fine-tune only the classifier on a small validation set. 

\vspace{-0.4cm}
\subsubsection{Experiments and Results:} On
\uline{ISIC,} 
the proposed model was successful in extracting invariant representations, evidenced both by the confused DANN domain classifier 
(Figure~\ref{fig:domain_accs}B) 
and the substantially reduced SOM purity (Fig~\ref{fig:som_plots}E). 
After fine-tuning this invariant representation on a small balanced validation set as used in DFR, the proposed approach 
achieved the highest worst-performance among all models considered (Table \ref{tab:all_methods_results}). 
Its performance on the bias-conflicting subgroup was better than GDRO with group adjustment, and its average performance on par. 
Note that while DFR exhibited a stronger average performance on the bias-conflicting subgroup, its worst-group performance was much lower. 
In comparison, the proposed method was the most successful in reducing subgroup disparity (last two columns in Table \ref{tab:all_methods_results}) with strong average performance.

\begin{table*}[t] \centering
\resizebox{\linewidth}{!}{ 
\begin{tabular}{@{\extracolsep{4pt}}lcccccccc@{\extracolsep{4pt}}}
\toprule
    & Average & Worst  & MN-12 & MN-56 & NN-12 & NN-56 \\

\midrule
ERM %
    & $ \cellcolor{shadecolor}0.810 \pm 0.000$& $ 0.617 \pm 0.040$
    & $ 0.878 \pm 0.008$ & $ 0.810 \pm 0.018 $ & $ 0.777  \pm 0.010 $& $ \cellcolor{shadecolor}0.855 \pm 0.005$  \\

 Importance weighting%

    & $ 0.803 \pm 0.015$ & $ 0.673 \pm 0.035$
    & $ 0.878 \pm 0.010$ & $ 0.853 \pm 0.008 $ & $ 0.780  \pm 0.001 $ & $ 0.853 \pm 0.010$ \\

    DANN %
    & $ 0.787 \pm 0.006$ & $ 0.607 \pm 0.015$ 
    & $ 0.857 \pm 0.010$ & $ 0.758 \pm 0.014 $ & $ 0.785  \pm 0.005 $  & $ 0.800 \pm 0.015$ \\

 DFR%
    & $ 0.790 \pm 0.010$ & $ 0.660 \pm 0.036$
    & $ 0.863 \pm 0.003$ & $ 0.813 \pm 0.020 $ & $ 0.778  \pm 0.008 $ & $ 0.828 \pm 0.010$ \\

GDRO %
    & $ 0.800 \pm 0.010$ & $ 0.640 \pm 0.030$
    & $ 0.868 \pm 0.008$ & $ 0.848 \pm 0.055 $ & $ 0.788  \pm 0.012 $  & $ 0.825 \pm 0.020$ \\


GDRO with group adjustment 
    & $ 0.780 \pm 0.010$ & $ 0.647 \pm 0.006$
    & $ 0.852 \pm 0.008$ & $ \cellcolor{shadecolor}0.877 \pm 0.016 $ & $ 0.782  \pm 0.015 $ & $ 0.788 \pm 0.022$ \\

Proposed method %
    & $ \cellcolor{shadecolor}0.810 \pm 0.000$ & $ \cellcolor{shadecolor}0.680 \pm 0.026$ 
    & $ \cellcolor{shadecolor}0.883 \pm 0.003$ & $ 0.837 \pm 0.003 $ & $ \cellcolor{shadecolor}0.803  \pm 0.003 $ & $ 0.830 \pm 0.009$ \\

\bottomrule
\end{tabular}
}
\caption{\label{tab:all_methods_results_fitz}
AUC of benign, malignant, and non-neoplastic skin lesion classification on Fitzpatrick. MN-12/-56: malignant with skin type 1-2 / 5-6. NN-12/56: non-neoplastic with skin type 1-2 / 5-6. MN-56 and NN-12 are the primary bias-conflicting subgroups.
}
\vspace{-.5cm}
\end{table*}

\begin{table*}[tb] \centering
\resizebox{\linewidth}{!}{ 
\begin{tabular}{@{\extracolsep{4pt}}lcccccccc@{\extracolsep{4pt}}}
\toprule
    & $\Delta_{mn}$ & $\Delta_{nn}$ & $\Delta_{bg}$ & $\Delta_{avg}$ \\

\midrule
ERM %
    & $ 0.140 \pm 0.026 $ & $ 0.083 \pm 0.015 $ & $0.223 \pm 0.032 $ & $0.149 \pm 0.020 $ \\

Importance Weighting  %
    & $ 0.073 \pm 0.021 $ & $ 0.080 \pm 0.020 $ & $0.153 \pm 0.006 $ & $0.102 \pm 0.010 $ \\

DANN %
    & $ 0.250 \pm 0.036 $ & $ 0.040 \pm 0.010 $ & $0.233 \pm 0.015 $ & $0.174 \pm 0.018 $ \\

DFR %
    & $ 0.157 \pm 0.040 $ & $ 0.053 \pm 0.006 $ & $0.150 \pm 0.030 $ & $0.120 \pm 0.021 $ \\

GDRO %
    & $ 0.090 \pm 0.087 $ & $ 0.053 \pm 0.025 $ & $0.197 \pm 0.055 $ & $0.113 \pm 0.055 $ \\

GDRO with group adjustment%
    & $ \cellcolor{shadecolor}0.053 \pm 0.015 $ & $ \cellcolor{shadecolor}0.030 \pm 0.010 $ & $0.163 \pm 0.021 $ & $\mathbf{0.082 \pm 0.004} $ \\

Proposed method %
    & $ 0.090 \pm 0.010 $ & $ \mathbf{0.037 \pm 0.006}$ & 
    $ \cellcolor{shadecolor}0.117 \pm 0.031 $ & $\cellcolor{shadecolor}0.081 \pm 0.008 $ \\

\bottomrule
\end{tabular}
}
\caption{\label{tab:all_methods_results_fitz_delta}
Difference in AUCs between best- and worst-performing subgroups in each class, from left to right: malignant, non-neoplastic, benign, and averaged.
}
\vspace{-.4cm}
\end{table*}

Evidence of improved invariant representations was similar on \uline{Fitzpartrick} (results shown in supplemental materials). 

As summarized in Tables \ref{tab:all_methods_results_fitz}-\ref{tab:all_methods_results_fitz_delta}, 
compared to reweighting methods, the fine-tuned classifier as proposed was the only one that was able to reduce the bias against NN-12, and delivered the best worst-group as well as average performance. 
Overall, its performance in removing subgroup disparity was on par with GDRO with group adjustment (Table \ref{tab:all_methods_results_fitz_delta}) while delivering significantly higher average and worst-group AUCs (Table \ref{tab:all_methods_results_fitz}).


\vspace{-0.4cm}
\subsubsection{Conclusions and Discussion:} 

We presented an evaluation study that derived important new insights into 
the working mechanisms and limitations of DRO and 
invariant representation learning to address spurious correlation caused by underrepresentation. 
The findings motivated us to present 
a novel approach that leverages robust optimization 
to facilitate the learning of invariant representations at the presence of spurious correlations. 
Finetuned classifiers utilizing such representation 
demonstrated an improved ability to reduce subgroup performance disparity while maintaining high average and worst-group performance. 
Future investigations will include a broader spectrum of approaches including those utilizing data augmentation, 
as well as extending to a wider range of medical image datasets exploring potential hidden biases. 

\section{Acknowledgments}

This work is supported by the National Institute of Nursing Research (NINR) of the National Institutes of Health (NIH) under Award Number R01NR018301.

%
%
%
%
\bibliographystyle{splncs04}
\bibliography{ref}
\end{document}


\title{Distributionally Robust Optimization and Invariant Representation Learning for Addressing Subgroup Underrepresentation:\\ Mechanisms and Limitations: \\  \large Supplemental Document}
%
%
\author{Nilesh Kumar$^*$, Ruby Shrestha$^*$, Zhiyuan Li \and Linwei Wang}
\institute{Rochester Institue of Technology, NY. USA}
\vspace{.3cm}  

%
%
%
\maketitle              
\vspace{-.3cm}

\section{SOM Purity Values for ISIC and Fitzpatrick Datasets}
\subsection{ISIC Dataset}
\vspace{-1.2em}
\begin{table*}[!h] \centering
\resizebox{\linewidth}{!}{ 
\begin{tabular}{@{\extracolsep{4pt}}lcccccccc@{\extracolsep{4pt}}}
\toprule
    & Benign, No Bandage & Benign, Bandage & Malignant, No Bandage & Malignant, Bandage & Average \\

\midrule
ERM %
   & $ 0.763\pm0.024$ & $0.945\pm0.007$ & $0.823\pm0.024$ & $0.730\pm0.014$ & $0.815\pm0.010$\\

GDRO 
& $ 0.768\pm0.088$ & $0.953\pm0.010$ & $0.763\pm0.052$ & $0.710\pm0.090$&$0.798\pm0.015$\\

GDRO with group adjustment %
& $ 0.803\pm0.061$ & $0.973\pm0.000$ & $\cellcolor{shadecolor}0.725\pm0.049$ & $0.743\pm0.004$&$0.811\pm0.029$\\
    
DANN %
& $ 0.802\pm0.040$ & $0.940\pm0.057$ & $0.775\pm0.106$ & $0.746\pm0.019$&$0.816\pm0.007$\\

Proposed method %
& $ \cellcolor{shadecolor}0.685\pm0.021$ & $\cellcolor{shadecolor}0.830\pm0.028$ & $\mathbf{0.755\pm0.007}$ & $\cellcolor{shadecolor}0.620\pm0.014$&$\cellcolor{shadecolor}0.723\pm0.007$\\

\bottomrule
\end{tabular}}
\caption{\label{tab:purity_isic}
SOM purities for subgroups in ISIC dataset. The proposed method resulted in the least biases in latent representations on average and for almost all the subgroups with either the lowest or the second-lowest SOM purity measures. Best performance is shaded gray, with bolded performance coming closely behind.}

\end{table*}
\vspace{-2em}
\subsection{Fitzpatrick Dataset}
\vspace{-1.2em}
\begin{table*}[!h] \centering
\resizebox{\linewidth}{!}{ 
\begin{tabular}{@{\extracolsep{4pt}}lcccccccc@{\extracolsep{4pt}}}
\toprule

    & Malignant-1,2 & Malignant-5,6 & Non-neoplastic-1,2 & Non-neoplastic-5,6 & Average \\

\midrule
ERM %
&$0.615\pm0.031$&$0.500\pm0.000$&$0.915\pm0.011$&$0.624\pm0.034$&$0.664\pm0.019$\\

GDRO 
&$\cellcolor{shadecolor}0.555\pm0.008$&$0.459\pm0.059$&$0.902\pm0.013$&$0.648\pm0.000$&$0.641\pm0.016$\\

GDRO with group adjustment %
&$0.588\pm0.008$&$0.500\pm0.000$&$\cellcolor{shadecolor}0.895\pm0.007$&$0.673\pm0.025$&$0.664\pm0.006$\\
    
DANN %
&$0.593\pm0.062$&$0.459\pm0.059$&$0.908\pm0.015$&$0.597\pm0.005$&$0.639\pm0.004$\\

Proposed method %
&$\mathbf{0.571\pm0.016}$&$\cellcolor{shadecolor}0.271\pm0.088$&$\mathbf{0.899\pm0.012}$&$\cellcolor{shadecolor}0.583\pm0.015$&$\cellcolor{shadecolor}0.581\pm0.027$\\

\bottomrule
\end{tabular}}
\caption{\label{tab:purity_fitz}
SOM purities for subgroups in Fitzpatrick dataset. In Fitzpatrick dataset as well, the proposed method resulted
in the least biases in latent representations on average and for almost all the subgroups
with either the lowest or the second-lowest SOM purity measures.}
\end{table*}
\vspace{-2.5em}
\begin{figure}
    \centering
    \includegraphics[width=1\textwidth,scale=0.65]{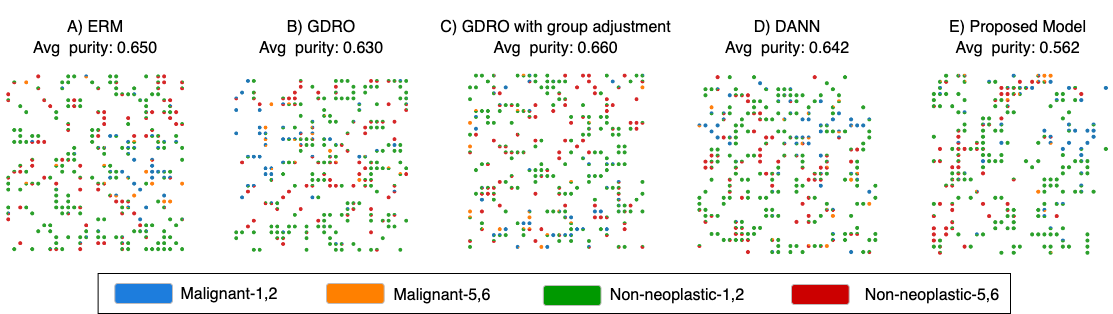}
    \caption{SOM plots and averaged subgroup purities for Fitzpatrick dataset. Malignant-1,2/-5,6 and Non-neoplastic-1,2/-5,6 are the subgroups where the bias primarily exists.
    }
    \label{fig:som_plots}
\end{figure}
\newpage
\section{Complete Performance on Fitzpatrick Dataset}
\vspace{-2.5em}
\begin{table*}[!h] \centering
\resizebox{\linewidth}{!}{ 
\begin{tabular}{@{\extracolsep{4pt}}lcccccccc@{\extracolsep{4pt}}}
\toprule
 \diagbox{Methods}{Benign Skin Types}  & 1 & 2 & 3 & 4 & 5 & 6 & $\Delta_{best-worst}$\\

\midrule
ERM %
   & $\cellcolor{shadecolor}0.770 \pm 0.020$ & $\cellcolor{shadecolor}0.723 \pm 0.015$ & $0.780 \pm 0.026$ & $0.757 \pm 0.040$ & $\mathbf{0.837 \pm 0.049}$ & $0.617 \pm 0.040$ & $0.223 \pm 0.032$ \\

 Importance weighting%

    & $0.747 \pm 0.031$ & $0.673 \pm 0.035$ & $0.767 \pm 0.029$ & $0.747 \pm 0.029$ & $0.827 \pm 0.031$ & $0.697 \pm 0.015$ & $0.153 \pm 0.006$ \\

DANN %
    & $0.710 \pm 0.010$ & $0.680 \pm 0.010$ & $0.743 \pm 0.006$ & $\cellcolor{shadecolor}0.763 \pm 0.012$ & $\cellcolor{shadecolor}0.840 \pm 0.010$ & $0.607 \pm 0.015$ & $0.233 \pm 0.015$ \\

DFR%
    & $0.727 \pm 0.012$ & $0.673 \pm 0.015$ & $0.757 \pm 0.025$ & $0.757 \pm 0.032$ & $0.810 \pm 0.010$ & $0.687 \pm 0.059$ & $0.137 \pm 0.015$\\

GDRO %
    & $0.733 \pm 0.032$ & $0.673 \pm 0.035$ & $0.777 \pm 0.015$ & $0.750 \pm 0.020$ & $0.837 \pm 0.031$ & $0.700 \pm 0.082$ & $0.197 \pm 0.055$ \\


GDRO with group adjustment 
    & $0.703 \pm 0.015$ & $0.647 \pm 0.006$ & $0.723 \pm 0.021$ & $0.710 \pm 0.017$ & $0.810 \pm 0.020$ & $\cellcolor{shadecolor}0.710 \pm 0.036$ & $0.163 \pm 0.021$\\

Proposed method %
    & $0.753 \pm 0.006$ & $0.697 \pm 0.006$ & $\cellcolor{shadecolor}0.797 \pm 0.006$ & $0.753 \pm 0.006$ & $0.787 \pm 0.006$ & $0.690 \pm 0.035$ & $\cellcolor{shadecolor}0.117 \pm 0.031$\\

\bottomrule
\end{tabular}
}
\caption{\label{tab:all_methods_results_fitz}
AUC values of individual skin types in the Benign category. The proposed method performed the best in removing the overall subgroup disparity. Best performance is shaded gray, with bolded performance coming closely behind.
}
\vspace{-.5cm}
\end{table*}
\vspace{-2.5em}
\begin{table*}[!h] \centering
\resizebox{\linewidth}{!}{ 
\begin{tabular}{@{\extracolsep{4pt}}lcccccccc@{\extracolsep{4pt}}}
\toprule
 \diagbox{Methods}{Malignant Skin Types}  & 1 & 2 & 3 & 4 & 5 & 6 & $\Delta_{best-worst}$\\

\midrule
ERM %
& $0.887 \pm 0.012$ & $\mathbf{0.870 \pm 0.010}$ & $0.867 \pm 0.021$ & $0.830 \pm 0.010$ & $0.873 \pm 0.006$ & $0.747 \pm 0.032$ & $0.140 \pm 0.026$\\

 Importance weighting%
& $0.893 \pm 0.015$ & $0.863 \pm 0.006$ & $\cellcolor{shadecolor}0.897 \pm 0.006$ & $\cellcolor{shadecolor}0.873 \pm 0.015$ & $0.880 \pm 0.010$ & $0.827 \pm 0.015$ & $0.073 \pm 0.021$\\

DANN %
& $0.863 \pm 0.012$ & $0.850 \pm 0.010$ & $0.840 \pm 0.000$ & $0.783 \pm 0.015$ & $0.883 \pm 0.006$ & $0.633 \pm 0.032$ & $0.250 \pm 0.036$\\

DFR%
& $0.877 \pm 0.006$ & $0.850 \pm 0.010$ & $0.870 \pm 0.017$ & $0.813 \pm 0.006$ & $\cellcolor{shadecolor}0.890 \pm 0.010$ & $0.737 \pm 0.038$ & $0.153 \pm 0.046$\\

GDRO %
& $0.877 \pm 0.006$ & $0.860 \pm 0.010$ & $0.883 \pm 0.021$ & $0.850 \pm 0.017$ & $0.887 \pm 0.006$ & $0.810 \pm 0.104$ & $0.090 \pm 0.087$\\


GDRO with group adjustment 
& $0.863 \pm 0.006$ & $0.840 \pm 0.010$ & $0.893 \pm 0.006$ & $0.850 \pm 0.000$ & $0.887 \pm 0.006$ & $\cellcolor{shadecolor}0.867 \pm 0.029$ & $\cellcolor{shadecolor}0.053 \pm 0.015$\\

Proposed method %
& $\cellcolor{shadecolor}0.897 \pm 0.006$ & $\cellcolor{shadecolor}0.870 \pm 0.000$ & $0.883 \pm 0.006$ & $0.853 \pm 0.012$ & $0.867 \pm 0.006$ & $0.807 \pm 0.006$ & $0.090 \pm 0.010$\\

\bottomrule
\end{tabular}
}
\caption{\label{tab:all_methods_results_fitz}
AUC values of individual skin types in the Malignant category. The proposed method delivered improvement over DANN in reducing the overall subgroup disparity and it was the third-best method in terms of $\Delta_{best-worst}$.
}
\vspace{-.5cm}
\end{table*}
\vspace{-2.5em}
\begin{table*}[!h] \centering
\resizebox{\linewidth}{!}{ 
\begin{tabular}{@{\extracolsep{4pt}}lcccccccc@{\extracolsep{4pt}}}
\toprule
 \diagbox{Methods}{Non-neoplastic Skin Types}  & 1 & 2 & 3 & 4 & 5 & 6 & $\Delta_{best-worst}$\\

\midrule
ERM %
& $0.777 \pm 0.015$ & $0.777 \pm 0.006$ & $\cellcolor{shadecolor}0.810 \pm 0.010$ & $\mathbf{0.837 \pm 0.012}$ & $\cellcolor{shadecolor}0.857 \pm 0.006$ & $0.853 \pm 0.006$ & $0.083 \pm 0.015$\\

 Importance weighting%
& $0.780 \pm 0.010$ & $0.780 \pm 0.010$ & $0.800 \pm 0.010$ & $0.833 \pm 0.015$ & $0.847 \pm 0.006$ & $\cellcolor{shadecolor}0.860 \pm 0.017$ & $0.080 \pm 0.020$\\

DANN %
& $0.790 \pm 0.010$ & $0.780 \pm 0.000$ & $0.803 \pm 0.006$ & $0.820 \pm 0.010$ & $0.807 \pm 0.015$ & $0.793 \pm 0.015$ & $0.040 \pm 0.010$\\

DFR%
& $0.780 \pm 0.010$ & $0.777 \pm 0.006$ & $0.800 \pm 0.010$ & $0.823 \pm 0.006$ & $0.830 \pm 0.010$ & $0.827 \pm 0.012$ & $0.053 \pm 0.006$\\

GDRO %
& $0.793 \pm 0.015$ & $0.783 \pm 0.012$ & $0.797 \pm 0.012$ & $0.813 \pm 0.015$ & $0.817 \pm 0.031$ & $0.833 \pm 0.012$ & $0.053 \pm 0.025$\\


GDRO with group adjustment 
& $0.787 \pm 0.015$ & $0.777 \pm 0.015$ & $0.773 \pm 0.012$ & $0.793 \pm 0.006$ & $0.780 \pm 0.020$ & $0.797 \pm 0.025$ & $\cellcolor{shadecolor}0.030 \pm 0.010$\\

Proposed method %
& $\cellcolor{shadecolor}0.807 \pm 0.006$ & $\cellcolor{shadecolor}0.800 \pm 0.000$ & $0.800 \pm 0.000$ & $\cellcolor{shadecolor}0.837 \pm 0.006$ & $0.827 \pm 0.006$ & $0.833 \pm 0.012$ & $\mathbf{0.037 \pm 0.006}$\\

\bottomrule
\end{tabular}
}
\caption{\label{tab:all_methods_results_fitz}
AUC values of individual skin types in the Non-neoplastic category. The proposed method was the only one that decreased the bias against Non-neoplastic skin types 1 and 2. Also, its performance in removing overall subgroup disparity was on par with GDRO with group adjustment and was better than DANN.
}
\vspace{-.5cm}
\end{table*}